\documentclass[conference]{IEEEtran}
\IEEEoverridecommandlockouts
\usepackage{cite}
\usepackage{amsmath,amssymb,amsfonts}
\usepackage{algorithmic}
\usepackage{graphicx}
\usepackage{textcomp}
\usepackage{float}
\usepackage{subfig}
\usepackage{color}
\usepackage[pagebackref,breaklinks,colorlinks]{hyperref}
\usepackage[capitalize]{cleveref}
\usepackage{booktabs}
\usepackage[table]{xcolor}
\usepackage{bm}
\usepackage{multirow}
\newcommand{\tablestyle}[2]{\setlength{\tabcolsep}{#1}\renewcommand{\arraystretch}{#2}\centering\footnotesize}
\def\BibTeX{{\rm B\kern-.05em{\sc i\kern-.025em b}\kern-.08em
    T\kern-.1667em\lower.7ex\hbox{E}\kern-.125emX}}
\begin{document}

\title{What a Whole Slide Image Can Tell? Subtype-guided Masked Transformer for Pathological Image Captioning\\
}


\author{
\IEEEauthorblockN{\quad\quad\quad Wenkang Qin$^{*}$\quad\quad\quad }
\IEEEauthorblockA{\textit{\quad\quad\quad College of Engineering\quad\quad\quad } \\
\textit{\quad\quad\quad Peking University\quad\quad\quad }\\
\quad\quad\quad Beijing, China\quad\quad\quad  \\
\quad\quad\quad qinwk@stu.pku.edu.cn\quad\quad\quad }
\and
\IEEEauthorblockN{\quad\quad\quad\quad Rui Xu$^{*}$\quad\quad\quad\quad }
\IEEEauthorblockA{\textit{\quad\quad\quad\quad College of Engineering\quad\quad\quad\quad } \\
\textit{\quad\quad\quad\quad Peking University\quad\quad\quad\quad }\\
\quad\quad\quad\quad Beijing, China\quad\quad\quad\quad  \\
\quad\quad\quad\quad xurui@stu.pku.edu.cn\quad\quad\quad\quad }
\and
\IEEEauthorblockN{\quad\quad\quad Peixiang Huang$^{*}$\quad\quad\quad }
\IEEEauthorblockA{\textit{\quad\quad\quad College of Engineering\quad\quad\quad } \\
\textit{\quad\quad\quad Peking University\quad\quad\quad }\\
\quad\quad\quad Beijing, China\quad\quad\quad  \\
\quad\quad\quad huangpx@stu.pku.edu.cn\quad\quad\quad }
\and
\IEEEauthorblockN{\quad\quad\quad\quad Xiaomin Wu\quad\quad\quad }
\IEEEauthorblockA{\textit{\quad\quad\quad\quad College of Engineering\quad\quad\quad } \\
\textit{\quad\quad\quad\quad Peking University\quad\quad\quad }\\
\quad\quad\quad\quad Beijing, China\quad\quad\quad  \\
\quad\quad\quad\quad xiaominw@stu.pku.edu.cn\quad\quad\quad }
\and
\IEEEauthorblockN{\quad\quad\quad\quad Heyu Zhang\quad\quad\quad }
\IEEEauthorblockA{\textit{\quad\quad\quad\quad College of Engineering\quad\quad\quad } \\
\textit{\quad\quad\quad\quad Peking University\quad\quad\quad }\\
\quad\quad\quad\quad Beijing, China\quad\quad\quad  \\
\quad\quad\quad\quad heyu@stu.pku.edu.cn\quad\quad\quad }
\and

\IEEEauthorblockN{\quad\quad\quad\quad Lin Luo$^{\dagger}$\quad\quad\quad}
\IEEEauthorblockA{\textit{\quad\quad\quad\quad College of Engineering\quad\quad\quad} \\
\textit{\quad\quad\quad\quad Peking University\quad\quad\quad}\\
\quad\quad\quad\quad Beijing, China\quad\quad\quad \\
\quad\quad\quad\quad luol@pku.edu.cn\quad\quad\quad}
\thanks{$*$ These authors contributed equally to this work.}
\thanks{$\dagger$ Corresponding author.}
}

\maketitle

\begin{abstract}
Pathological captioning of Whole Slide Images (WSIs), though is essential in computer-aided pathological diagnosis, has rarely been studied due to the limitations in datasets and model training efficacy.
In this paper, we propose a new paradigm Subtype-guided Masked Transformer (SGMT) for pathological captioning based on Transformers, which treats a WSI as a sequence of sparse patches and generates an overall caption sentence from the sequence.
An accompanying subtype prediction is introduced into SGMT to guide the training process and enhance the captioning accuracy.
We also present an Asymmetric Masked Mechansim approach to tackle the large size constraint of pathological image captioning, where the numbers of sequencing patches in SGMT are sampled differently in the training and inferring phases, respectively.
Experiments on the PatchGastricADC22 dataset demonstrate that our approach effectively adapts to the task with a transformer-based model and achieves superior performance than traditional RNN-based methods.
Our codes are to be made available for further research and development.
\end{abstract}

\section{Introduction}
Histopathology is a cornerstone of cancer diagnosis, treatment planning, and prognostication\cite{rorke1997pathologic,zhang2019pathologist}. The microscopic examination of tissue specimens by pathologists enables identification of morphological abnormalities and helps to guide clinical decisions. The emergence of digital pathology has revolutionized the field by digitally scanning tissue specimens into high-resolution images which can be analyzed and shared electronically\cite{melo2020whole,qin2022pathtr}, and thus bring benefits in diagnostic accuracy, remote diagnosis, and enhanced quality control, etc\cite{rodriguez2022artificial}.

Deep-learning algorithms emerge to achieve automatic analysis of digital pathology Whole Slide Images (WSIs) with high efficiency and accuracy \cite{lu2021data}. The models need to adapt to large image size, data variability and complexity. More importantly, the analysis results requires pathologists to interpret and conclude into pathological reports with their expertise\cite{barker2016automated}.

Automated pathological captioning, which involves generating textual descriptions of WSIs, is the task that aids in the analysis and interpretation of pathological images\cite{patchgastic} for human understanding. Pathological captioning can provide concise descriptions of the typical diagnostic features of a WSI, and guide pathologists to focus on the regions of interest\cite{beddiar2022automatic}. It is essential for computer-aided diagnosis to ensure the accuracy and reliability of pathological analysis through integration of computational methods\cite{xiong2019reinforced}.

Despite the potential benefits, automated pathological captioning has not been fully investigated due to the lack of high quality datasets\cite{gamper2021multiple,zhang2022pathnarratives,he2020pathvqa}, and the training of captioning models on large-size and variable pathological WSI images. Since last year, some researchers started to use recurrent neural networks (RNNs)\cite{graves2012long},learning for pathological captioning\cite{gamper2021multiple,patchgastic}, but they are limited in capturing long-range dependencies and potential risk of vanishing gradients.
Moreover, traditional natural image captioning methods which utilize dense grid features\cite{zhang2021rstnet} and sparse region of interest (ROI) features\cite{shao2023textual,peng2022fast} from Region Proposal Network (RPN)\cite{ren2015faster} are not directly applicable for pathological images, because the lesion attention regions in a WSI are too distributed and complex for RPN to discover for diagnosis, while dense grid features may result in too long sequence lengths as captions, and thus making the model costly in time and memory consumption and prone to overfitting or sensitive to noise.
\begin{figure*}[t]
\centering
\includegraphics[width=\linewidth]{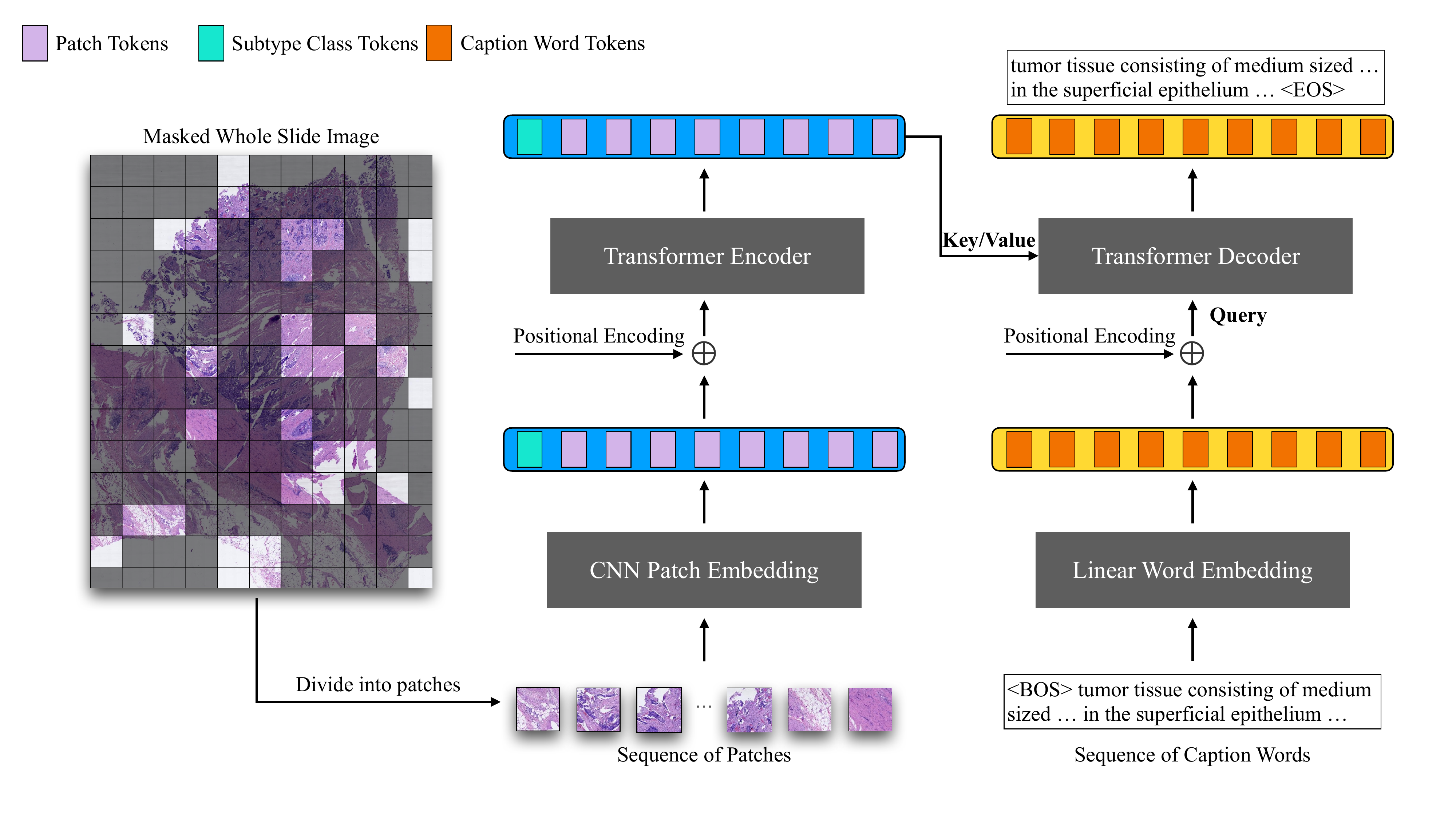}
\caption{The proposed Subtype-guided Masked Transformer (SGMT) for pathological image captioning.}
\label{fig:SGMT}
\end{figure*}
To address these challenges, we propose a new paradigm \emph{Subtype-guided Masked Transformer} (SGMT) for pathological captioning that leverages the power of transformer-based models\cite{vaswani2017attention,brown2020language} for their excellent context-association capability. In SGMT, a sparse-grid masking module is introduced to suggest candidate features for pathological caption generation in order to control the captions to be comprehensive but within a reasonable length, where the masking mechanism of training and inferring are asymmetric. We then design a subtype-guided prediction module to guide the model to generate more informative and accurate captions out of the heterogeneity and complexity of pathological images.
A mechanism of random sampling and voting strategy is also added to optimize and ensure the stability of caption outputs.
The proposed SGMT paradigm provides a flexible framework to generate comprehensive and concise captions for large-size pathological images. It is flexible and can be extended to other histopathology-related tasks such as visual question answering by adjusting certain modules, e.g. the masking mechanism and voting strategy.

We evaluate the approach on the PatchGastricADC22 dataset \cite{patchgastic}, which contains over 262K patches of gastric adenocarcinoma.
Results show that SGMT outperforms traditional RNN-based approaches in terms of both quantitative metrics and visual quality of the generated captions.
We also conduct ablation studies to show the contribution of each component, which proves the benefits of incorporating the subtype-guided prediction module.
Besides providing accurate and efficient pathological captioning, SGMT can be extended to other histopathology-related tasks such as visual question answering. We expect this work will inspire the advancement of precise human-intelligence-collaborated diagnosis in digital pathology for better cancer care.

\section{Methods}

\subsection{Overview}
The overall architecture of our Subtype-guided Masked Transformer (SGMT) is illustrated in \cref{fig:SGMT}.
To tackle the problem of pathological captioning from WSIs, we first formulate it as a sequence-to-sequence task, where the input sequence contains a set of patches extracted from a WSI, and the output sequence is the corresponding caption describing the patches.
Unlike natural image captioning which uses sparse ROI features or dense grid features\cite{wang2022end,zhang2021rstnet,shao2023textual,peng2022fast}, a pathological WSI contains redundancy and the same information can present in many patches of the WSI, therefore SMTG adopts a sparse grid feature extraction approach with details described in \cref{chap:amt}.
The sequencing process is: given a WSI $I$, we first divide it into a set of patches and sample in a subsequence $P=(p_1,p_2,\ldots,p_n)$ with the same size, where $n$ is the total number of sampled patches.
Each patch $p_i$ is a square image with the size of $w_p \times w_p$.
We denote the caption of WSI $I$ as $y$. Then we can represent the WSI $I$ as a sampled sequence of patches and corresponding captions: $\{(p_1, p_2,\ldots, p_n), y\}$.

Next, we apply the Transformer \cite{vaswani2017attention} model to translate the sampled patch sequence to the caption sequence.
The Transformer model consists of an encoder and a decoder.
The encoder takes the patch sequence as input and transforms it into a sequence of high-level features.
The decoder then generates the caption sequence based on the encoded features and the previous generated caption.

The decoder generates the caption sequence auto-regressively.
At each time step $t$, it takes the embedded representation of the previous generated caption token $y_{t-1}$ and the context vector, which is the weighted sum of the feature vectors, as input.
The context vector is computed by a multi-head attention mechanism, which assigns different weights to different patches based on their relevance to the current generated caption token.
The decoder then generates the next caption token by applying a softmax function on the output of a feedforward neural network.

In addition to the standard Transformer \cite{vaswani2017attention}  model, we introduce a subtype prediction module that takes the encoded features as input and predicts the subtype label of the WSI.
This module is intended to guide the model to learn more discriminative features that can better capture the subtle differences among different subtypes of cancer. We will provide more details about this module in \cref{sec:subtype}.

\subsection{Subtype-guided Token}
\label{sec:subtype}
Different tumor subtypes are usually associated with rather different WSI captions, while the captions under the same subtype are similar to each other. The subtype information can be obtained from caption labeling as the equivalent of caption categories. Overall, Subtype classification can be predicted before the decoder and is potentially easier than captioning, which makes guiding captioning by subtype information possible.

We design the subtype-guidance mechanism as an accompanying task of WSI captioning. The subtype-guided token is initialized as a learnable embedding vector $e_{subtype} \in \mathbb{R}^{d}$, where $d$ is the embedding dimension.
It is then concatenated to the CNN-encoded patches and fed into the transformer encoder. During training, the subtype-guided token is optimized by minimizing the cross-entropy loss between the predicted and ground-truth subtypes.
Specifically, let $s_i$ be the ground-truth subtype for the patches, $\tilde{s_{i}}$ be the predicted subtype for the patches, and $\mathcal{L}_{subtype}$ be the cross-entropy loss, then we have:
\begin{equation}
	\mathcal{L}{subtype} = -\sum_{i=1}^{K} s_{i}\log(\tilde{s_{i}}),
\end{equation}
 where $K$ is the number of possible subtypes.
 The subtype-guided token, while being optimized during training, is used as the key and value in the self-attention mechanism of the transformer decoder to guide the generation of relevant and accurate captions.

\subsection{Asymmetric Masked Mechanism}
\label{chap:amt}
As mentioned before, dense-grid inputs have drawbacks:
1) Long training sequences implicating high time complexity and space complexity may make it difficult to train the model;
2) Information redundancy and noise in a WSI can cause model overfitting;
3) Long inference time in practical application.
The redundancy of information in a WSI allows the same information to be likely to appear in multiple patches, so it is possible to use sparse grid features.
We propose the Asymmetric Masked Mechanism (AMM) for sparse-grid feature extraction, which randomly samples different fixed number of patches during training and inference phases, respectively.
If the number of patches is less than the fixed number, all patches are used. This method helps to increase the model's generalization ability.
During inference, we also sample patches but with a larger upper limit to provide as much information as possible.

Formally, let $P = (p_1, p_2, \dots, p_n)$ be the set of patches, where $n$ is the number of patches, and $M$ be the upper limit for the number of patches to be sampled.
During training, we randomly sample $m$ patches from $P$ where $m = \min(M, n)$, and the remaining patches are masked out.
During inference, we sample $m'$ patches from $P$ where $m' = \min(\alpha M, n)$, and the remaining patches are masked out, where $\alpha > 1$ is a hyperparameter that controls the number of patches to be sampled during inference.
This sampling strategy helps to make the model more robust and generalizable.

We also note that during training, the patches are sampled independently for each training iteration to ensure diversity in the training data.
In addition, the patches are randomly permuted within each batch to further increase the diversity of the training data, which will be discussed in the experiments in \cref{sec:ablation}.

\subsection{Random Sampling and Voting Strategy}
The Asymmetric Masked Mechanism approach enables us to perform test-time data augmentation by \emph{performing multiple different data samplings} and augmentations on the input patches during inference.
Specifically, we randomly sample multiple sets of patches, and apply different augmentations such as random flipping and rotating to each set.
Then, we feed each set of patches into the model to obtain the corresponding captions. Finally, we use a voting scheme to combine the generated captions from different patch sets to produce the final output caption.
This approach helps to improve the robustness and accuracy of our model during inference, especially when the input patches have different degrees, flips or unmasked fields, and avoid inaccurate captions that may be caused by missing information during sampling.

In mathematical terms, let $P$ be the set of input patches during inference, $\mathcal{S}={S_1,S_2,\dots,S_k}$ be the set of $k$ random samplings from $P$, and $\mathcal{A}={A_1,A_2,\dots,A_k}$ be the set of $k$ augmented patch sets corresponding to $\mathcal{S}$. Each augmented patch set $A_i$ is obtained by applying random flipping, rotating and masking to the patches in $S_i$. Then, for each augmented patch set $A_i$, we feed it into the model to obtain the corresponding caption $y_i$. Finally, we combine the generated captions from all sets by voting:
\begin{equation}
	\hat{y}=\operatorname{vote}({y_i}_{i=1}^k)
\end{equation}
where $\operatorname{vote}$ returns the most frequent caption among ${y_i}_{i=1}^k$.

\begin{figure*}[htbp]
	\centering
	\subfloat[Patch limit during training]{
		\label{fig:patch_limit_training}
		\includegraphics[width=0.45\textwidth]{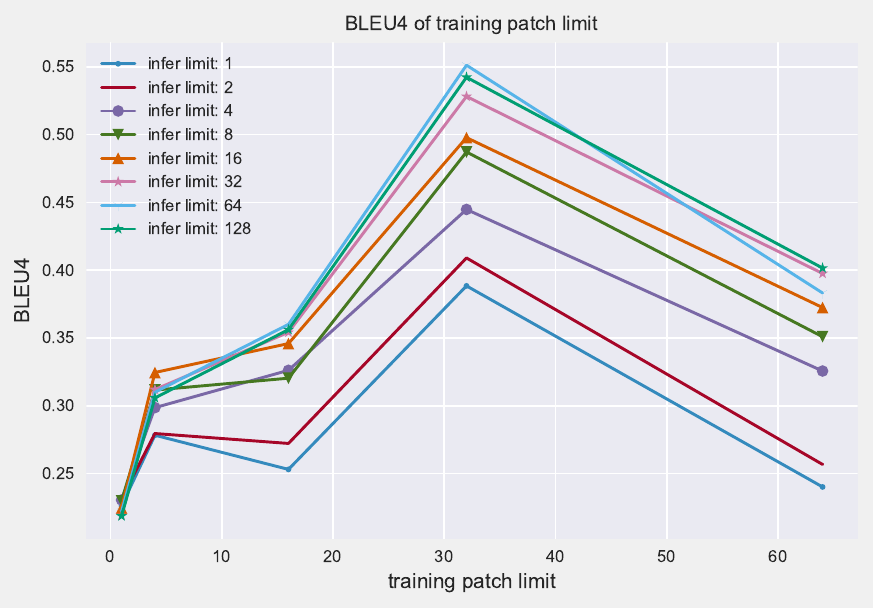}
	}
	\subfloat[Patch limit during testing]{
		\label{fig:patch_limit_testing}
		\includegraphics[width=0.45\textwidth]{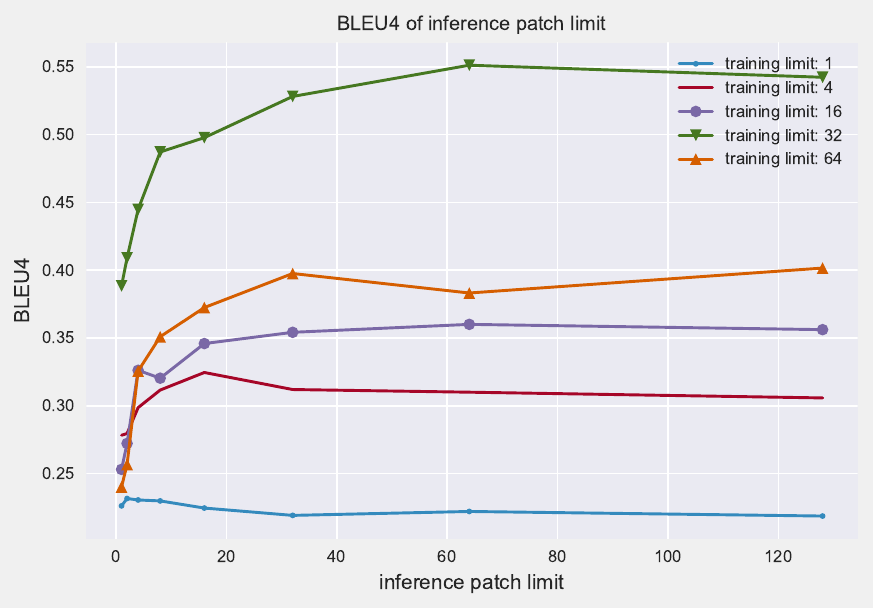}
	}
	\caption{Asymmetric Masked Mechanism performance with varying patch limits}
	\label{fig:patch_limit}
\end{figure*}
\begin{table*}
	\centering
 \resizebox{2.0\columnwidth}{!}{
    \tablestyle{20pt}{1.0}
	\begin{tabular}{ccccccc}
        \toprule
        ~ & BLEU-4 (\%) & CIDEr & ROUGE (\%) & METEOR (\%) \\
        \midrule
        LSTM (DenseNet121 pavg) & 27.2 & 1.52 & 46.95 & 28.21 \\
        LSTM (EfficientNetB3 pavg) & 28.3 & 2.31 & 49.73 & 30.54 \\
        \hline
        LSTM (DenseNet121 p3x3)  & 32.3 & 1.48 & 45.47 & 26.85 \\ 
        LSTM (EfficientNetB3 p3x3) & 32.4 & 1.62 & 46.49 & 27.75 \\ 
        \hline
        \rowcolor{gray!20}\textbf{Ours} & \textbf{55.11} & \textbf{4.83} & \textbf{69.68} & \textbf{43.17} \\
        \bottomrule
	\end{tabular}}
    \caption{Our SGMT compared with LSTM on PatchGastricADC22 \cite{patchgastic} dataset.}
    \label{table:main_results}
\end{table*}

\section{Experiments}

\subsection{Dataset}
We evaluate our SGMT on the PatchGastricADC22 dataset \cite{patchgastic}, which contains pairs of stomach adenocarcinoma endoscopic biopsy specimens patches and associated histopathological captions.. Specifically, patches from a slide correspond to the same caption extracted from the corresponding diagnostic report. The dataset consists of 262,777 patches extracted from 991 H\&E stained slides from distinct patients at a magnification of 20$\times$. Each patch with a pixel size of 300$\times$300 is mostly from regions containing lesions of nine gastric adenocarcinoma subtypes. The captions vocabulary consists of 277 words with a maximum sentence length of 50 words. The dataset is stratified by subtypes and divided in the ratio of 7:1:2 for training, validation and test.

\subsection{Experimental Details}
\noindent\textbf{Loss function}
To optimize the captioning task and the subtype classification task simultaneously, we used the cross-entropy function. The total loss function can be written as:
\begin{equation}
	\begin{aligned}
		\mathcal{L} &= \mathcal{L}_{caption} + \beta \mathcal{L}_{subtype} \\
		&=-\sum_{t=1}^T\log P(y_t|y_{<t}, I; \bm{\theta})-\beta\log P(s|I;\bm{\theta})
	\end{aligned}
\end{equation}
where $P(y_t|y_{<t}, I; \bm{\theta})$ is the probability of the $t$-th word in the caption given the previous words and the input image $I$, and $\beta$ is a hyperparameter to balance the two losses. $y_c$ is the ground-truth subtype label, and $P(s|I; \bm{\theta})$ is the predicted probability of the subtype given the input image $I$.

\noindent\textbf{Hyperparameters} We use the pre-trained ResNet-18 \cite{he2016deep} model trained on ImageNet as our image feature extractor and AdamW optimizer with a learning rate of 1e-5 and a weight decay of 1e-2. The batch size is set to 2. The model is trained for 40 epochs and the model weights from the last epoch are used for evaluation. Both the encoder and decoder of the Transformer are set to 6 layers. We implement our model in PyTorch and train it on a single NVIDIA Tesla V100 GPU.

For training the subtype-guided loss, we use a weight of $\beta$ = 1.0. We set the maximum patch number $M$ to 32 during training, and we test with maximum patch numbers $\alpha M$ of 64 during testing.

\begin{table*}
	\centering
 \resizebox{2.0\columnwidth}{!}{
    \tablestyle{20pt}{1.0}
	\begin{tabular}{lcccc}
		\toprule
		  ~ & BLEU-4 (\%) & CIDEr & ROUGE (\%) & METEOR (\%) \\
		\midrule
		Transformer with AMM & 48.84 & 4.32 & 67.32 & 41.21 \\
		+ subtype-guided token & 55.11 & 4.83 & 69.68 & 43.17 \\
		+ Sampling \& Voting & \textbf{55.80} & \textbf{4.99} & \textbf{70.78} & \textbf{43.92} \\
		\bottomrule
	\end{tabular}
 }
	\caption{Ablation study results on PatchGastricADC22 \cite{patchgastic} dataset.}
    \label{table:ablation_results}
\end{table*}
\begin{table}
\centering
 \resizebox{1.0\columnwidth}{!}{
    \tablestyle{1pt}{1.05}
\begin{tabular}{cccccc}
\toprule
Training Patch  & Inference Patch  & BLEU-4 (\%) & CIDEr & ROUGE (\%) & METEOR (\%) \\
\midrule
\multirow{8}{*}{1}  & 1   & 22.62  & 0.486 & 45.33 & 32.51  \\
                    & 2   & 23.16  & 0.501 & 45.77 & 32.95  \\
                    & 4   & 23.05  & 0.496 & 45.54 & 32.87  \\
                    & 8   & 22.98  & 0.505 & 45.49 & 32.78  \\
                    & 16  & 22.45  & 0.395 & 45.08 & 32.57  \\
                    & 32  & 21.91  & 0.343 & 44.85 & 32.35  \\
                    & 64  & 22.20  & 0.393 & 45.15 & 32.49  \\
                    & 128 & 21.87  & 0.352 & 44.86 & 32.29  \\ \hline
\multirow{8}{*}{4}  & 1   & 27.82  & 1.891 & 54.52 & 33.81  \\
                    & 2   & 27.95  & 1.823 & 55.28 & 34.18  \\
                    & 4   & 29.86  & 2.013 & 56.50 & 34.86  \\
                    & 8   & 31.15  & 2.143 & 57.89 & 35.34  \\
                    & 16  & 32.45  & 2.218 & 58.52 & 35.95  \\
                    & 32  & 31.19  & 2.125 & 58.44 & 35.71  \\
                    & 64  & 31.00  & 2.164 & 58.31 & 35.42  \\
                    & 128 & 30.57  & 2.072 & 58.06 & 35.44  \\ \hline
\multirow{8}{*}{16} & 1   & 25.31  & 1.810 & 49.79 & 29.03  \\
                    & 2   & 27.22  & 1.983 & 52.58 & 30.86  \\
                    & 4   & 32.61  & 2.462 & 58.10 & 33.86  \\
                    & 8   & 32.03  & 2.425 & 57.68 & 34.00  \\
                    & 16  & 34.58  & 2.634 & 59.98 & 35.31  \\
                    & 32  & 35.41  & 2.745 & 61.44 & 36.19  \\
                    & 64  & 36.00  & 2.761 & 61.81 & 36.48  \\
                    & 128 & 35.61  & 2.715 & 61.49 & 36.44  \\ \hline
\multirow{8}{*}{32} & 1   & 38.84  & 3.341 & 58.96 & 35.37  \\
                    & 2   & 40.90  & 3.481 & 59.88 & 36.12  \\
                    & 4   & 44.47  & 3.733 & 62.10 & 37.91  \\
                    & 8   & 48.71  & 4.225 & 66.84 & 41.00  \\
                    & 16  & 49.77  & 4.371 & 66.62 & 41.00  \\
                    & 32  & 52.80  & 4.656 & 68.88 & 42.52  \\
                    & 64  & 55.11  & 4.836 & 69.68 & 43.17  \\
                    & 128 & 54.22  & 4.778 & 69.50 & 42.99  \\ \hline
\multirow{8}{*}{64} & 1   & 24.01  & 1.683 & 48.49 & 29.39  \\
                    & 2   & 25.67  & 1.900 & 51.87 & 30.95  \\
                    & 4   & 32.56  & 2.502 & 56.59 & 34.25  \\
                    & 8   & 35.09  & 2.815 & 58.45 & 35.01  \\
                    & 16  & 37.24  & 2.972 & 58.15 & 35.57  \\
                    & 32  & 39.75  & 3.189 & 60.42 & 36.77  \\
                    & 64  & 38.31  & 3.165 & 59.43 & 35.72  \\
                    & 128 & 40.15  & 3.334 & 60.43 & 36.75 \\
\bottomrule
\end{tabular}}
\caption{Raw Metrics Data on Different Patch Limit.}
\label{table:raw_data}
\end{table}

\begin{figure}[h]
    \centering
    \includegraphics[width=\linewidth]{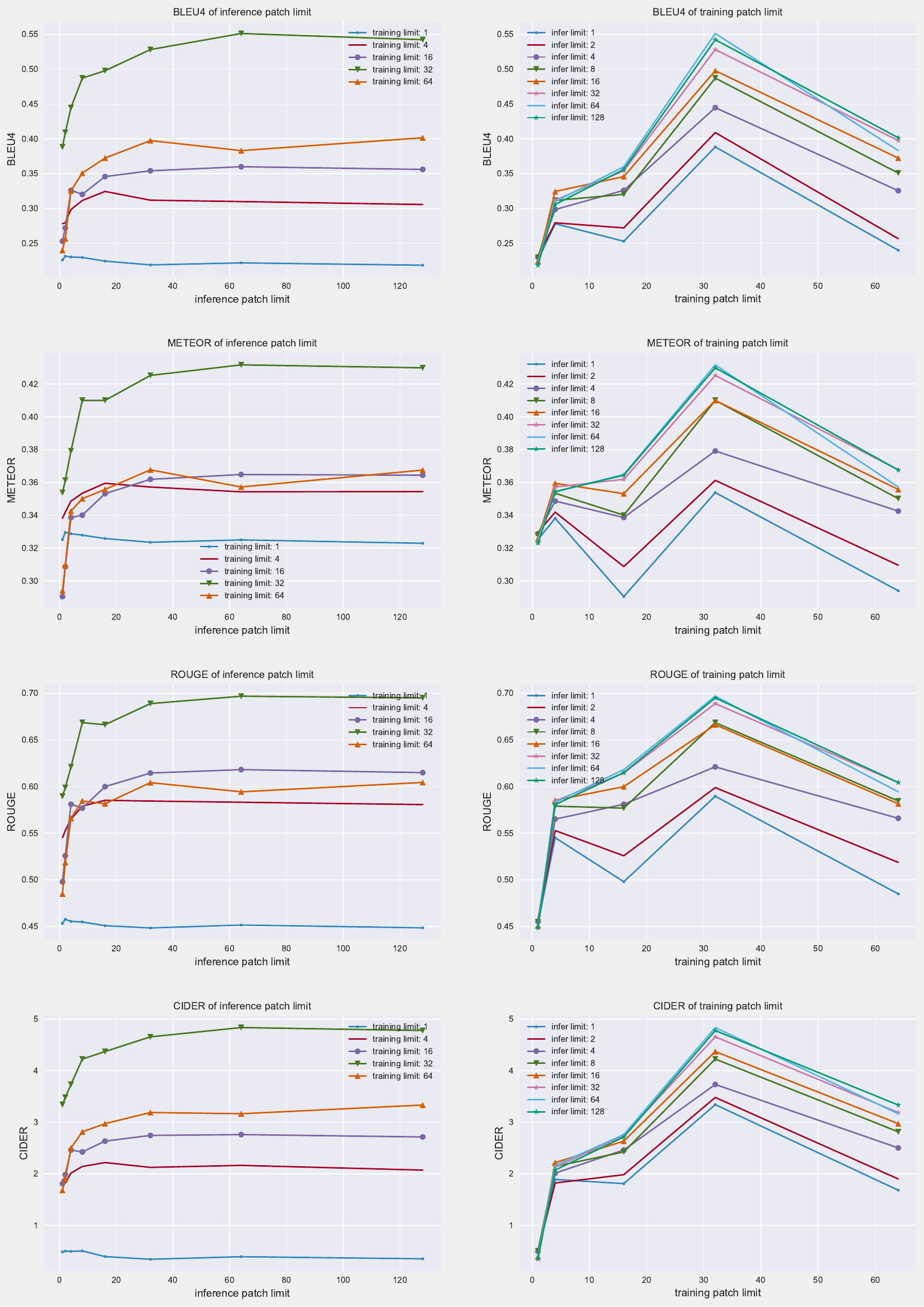}
    \caption{Metrics on Different Patch Limit.}
    \label{fig:raw_data}
\end{figure}

\subsection{Main Results}

We conducted experiments on the PatchGastricADC22 \cite{patchgastic} dataset to evaluate the performance of our proposed method, as well as to compare it with other existing methods. We use BLEU\cite{papineni2002bleu} and other commonly used metrics to evaluate the performance. \cref{table:main_results} shows the comparison results, where our method outperforms all the other methods significantly. Specifically, our proposed method (without random sampling and voting strategy) achieves a BLEU-4 score of 55.80, which is 70\% higher than the second-best method. These results demonstrate the effectiveness of our proposed approach for image captioning on the PatchGastricADC22 \cite{patchgastic} dataset.

\subsection{Ablation Study}
\label{sec:ablation}
In the ablation study, we compare the performance of our method with the baseline seq2seq model based on the vanilla Transformer \cite{vaswani2017attention} architecture with AMM, and then with the addition of subtype-guided token and our proposed random sampling and voting strategy (Sampling \& Voting in \cref{table:ablation_results}).

We evaluate our Asymmetric Masked Machanism approach with different patches limits during training and testing.
Specifically, we train models with different maximum numbers of patches in the input during training and test them with different maximum patch limits during inference.

As the ablation study results in \cref{table:ablation_results} show, our SGMT achieves significant improvements without subtype-guided token and random sampling and voting strategy in all evaluation metrics compared to all LSTM models in \cref{table:main_results}.
The addition of subtype-guided token also leads to further improvements, especially in terms of BLEU-4 score.
The proposed data augmentation strategy further boosts the performance of the model.
But even if the random sampling and voting strategy is not applied, the evaluation indicators will not drop sharply.
This shows that due to the redundancy existing in WSIs, using 64 as the sampling length for inference and cause hardly information loss.

The plot for Asymmetric Masked Mechanism in \cref{fig:patch_limit} shows the model's performance with varying patch limits during training and inference.
In \cref{fig:patch_limit_training}, the performance improves as the number of patches increases, but only up to a certain limit.
Beyond the limit of 32, the performance starts to degrade.
This suggests that too many patches may introduce noise that degrades the model's ability to learn useful information.
It also shows in \cref{fig:patch_limit_testing} that using more and more patches does not keep the model performance improving during inference. 
When using 128 inference patches, the model has been able to achieve the good enough performance.
Under the conditions of our experiment, that is, when the sampling ratio of WSIs is $20\times$ and the patch resolution is $300\times300$, the training sampling limit to 32 during training can improve the generalization ability of the model, and it only needs to use twice the number of patches during training to get good enough performance.
These settings not only ensure the performance of the model, but also reduce the time consumption of training and reasoning.

In the Table \ref{table:raw_data} and Fig. \ref{fig:raw_data}, we present the performance of the model under different constraints on the number of training and inference input patches. It can be seen that under different evaluation indicators, the model shows the redundancy of the input patch, and too many input patches are likely to cause overfitting. It can also be clearly seen that it is not the improvement of the input patch during inference, which does not bring continuous performance improvement, and less patch input is conducive to improving the inference speed of the model. It is the redundancy of WSI that inspired us to get such an approach, that is, less input ensures sample diversity during training, while ensuring inference speed.

\section{Conclusion and Discussion}

In this paper, we proposed a novel method called SGMT for WSI captioning based on the Transformers architecture. An accompanying task is introduced to generate pathological subtype-guided tokens for accurate captioning. We propose an asymmetric masked mechanism and a random sampling and voting strategy to further improve the performance. Experiments show that our method outperforms the state-of-the-art methods on the PatchGastricADC22 dataset.

PatchGastricADC22 dataset does not contain patch positions, thus the effect of patch positional encoding is not explored in this paper. In future work, we plan to investigate the use of position encoding in patch embeddings to better model the spatial information of the image. The proposed SGMT can be extended for other tasks in computational pathology, such as pathological images visual question answering.

\section*{Acknowledgment}
We thank Qiuchuan Liang (Beijing Haidian Kaiwen Academy, Beijing, China) for preprocessing data.






{
\bibliographystyle{IEEEtran}
\bibliography{IEEEabrvall}
}


\end{document}